\documentclass[runningheads,a4paper]{llncs}
\usepackage{amssymb}
\usepackage{amsmath}
\usepackage{graphicx}
\usepackage{subfigure}
\usepackage{url}
\urldef{\mailsa}\path|
{adityasharma101993}@gmail.com,
{prabhjot_kaur}@students.iitmandi.ac.in|
\urldef{\mailsb}\path|{aditya, arnav}@iitmandi.ac.in|
\newcommand{\keywords}[1]{\par\addvspace\baselineskip
	\noindent\keywordname\enspace\ignorespaces#1}

\usepackage{algorithm}
\usepackage{algpseudocode}
\usepackage{multirow}
\usepackage{array}
\usepackage{bm}
\usepackage{array}
\usepackage{makecell}
\usepackage{setspace} 

\begin{document}
	\mainmatter  
	\title{Learning to Decode 7T-like MR Image Reconstruction from 3T MR Images}
	
	\titlerunning{7T MR images from 3T MR images}
	%
	%
	\author{Aditya Sharma$^\ast$, Prabhjot Kaur$^\ast$, Aditya Nigam, Arnav Bhavsar}
	%
	
	

    \institute{School of Computing and Electrical Engineering\\ Indian Institute of Technology Mandi\\adityasharma101993@gmail.com, \;prabhjot\_kaur@students.iitmandi.ac.in,\;\{aditya, arnav@iitmandi.ac.in\}}

    %
	%
	
	\maketitle
	
	\begin{abstract}

Increasing demand for high field magnetic resonance (MR) scanner indicates the need for high-quality MR images for accurate medical diagnosis. However, cost constraints, instead, motivate a need for algorithms to enhance images from low field scanners. We propose an approach to process the given low field (3T) MR image slices to reconstruct the corresponding high field (7T-like) slices. Our framework involves a novel architecture of a merged convolutional autoencoder with a single encoder and multiple decoders. Specifically, we employ three decoders with random initializations, and the proposed training approach involves selection of a particular decoder in each weight-update iteration for back propagation. We demonstrate that the proposed algorithm outperforms some related contemporary methods in terms of performance and reconstruction time.{\let\thefootnote\relax\footnote{{$\ast$ The authors have contributed equally in this work. }}} 
		
	\end{abstract}

%

\section{Introduction}
Improvement of trade-off between spatial resolution and signal to noise ratio (SNR) in MR imaging motivates the research from the perspective of both hardware and signal processing.
As SNR increases monotonically with the strength of magnetic field, high-field MR scanners (7T, 11.5T) have been designed and are successful in providing higher SNR for the same resolution of images. However, the cost increases exponentially with the magnetic field strength. 
This leads to the lesser availability of high-field MR scanners across different hospitals and clinical labs and thus doesn't solve the problem in practice. The number of clinical 7T scanners in the world are just $\sim$40, as compared to $\sim$20000 3T scanners [5]. 
Thus, developing algorithms to enhance images from low-field (and low-cost) MR scanners, serve as an important alternative. 
Indeed, it has been shown that the signal processing techniques can improve the spatial resolution along with significant increment in the SNR~\cite{CanThey}. 


The problem to reconstruct the high-field like images from the low-field images is manifold and consists many sub-problems which include i) increase in resolution leading to enhancement of image details, 
ii) contrast improvement, and iii) increase in signal to noise ratio. Also, those approaches to address such problems are more feasible in clinical practices, which take less time.

One can address the above concerns by learning a highly non-linear mapping from the low field to high field MR images using exemplar low-field (LF) and high-field (HF) MR images. 
Considering this, Khosro et.al. in~\cite{TMI3T7T} attempted to construct 7T like MR images from 3T MR images using dictionaries defined in same space, which is estimated by hierarchical application of canonical correlation analysis (CCA), and 7T MR images are reconstructed using the dictionary defined for 7T MR images and the coefficient vector computed by representation of 3T MR images using dictionary of corresponding exemplary 3T MR images. As it tries to capture the non-linearity of the transformation, it performs better than the approaches which solely increase the resolution with SNR~\cite{MIMCS,YangWright}. However, the non-linearity of transformation is still approximated by linear operations and may have significant fitting errors by degree of the non-linearity. 

This is further addressed by the approaches defined in the popular framework of a neural network which can well approximate even the non-linear transformations~\cite{MICCAI2016,MICCAI2017}. In \cite{MICCAI2016} the reconstruction of 7T MR images is explored using convolutional neural network (CNN) network with a requirement of anatomical features. Reconstruction, as well as segmentation of the high-field MR images, is performed using a cascaded CNN given the 3T MR images and corresponding segmentation images at the input. Both these approaches divide the image volumes into 3D cubes and execute the algorithm with 3D CNN. Processing 3D cubes can help in reconstruction of local details and consistency in x,y, and z directions, but at the same time it may introduce block artifacts, and importantly, increases the time for reconstruct the test MR volumes. 


Considering the ill-posed nature of the problem, and a possibility of multiple good solutions, we propose a merged convolutional autoencoder with three decoders, along with a strategy to update the weights adaptively based on the performance of each decoder at every iteration. 
The final estimate of the HF image is obtained by averaging the reconstructed images from the three decoders. To make the algorithm better usable in clinical practices, we reduce the reconstruction time of test MR volumes, while achieving better reconstruction and segmentation of the high-field MR images, by processing 2D images, and removing the requirement of any anatomical/segmentation based features.


Thus, our contributions can be summarized as: a) architecture of convolutional autoencoder with multiple decoders. 
b) update criteria for the encoder weights on the basis of decoder performance. 
c) merge connections to enhance the reconstruction ability. d) demonstrating reconstruction and segmentation improvements along with significant reduction in reconstruction time as compared to the state-of-the-art approaches. {e) 
We demonstrate superior performance across a variety of quantitative metrics such as PSNR, SSIM, sharpness and edge width unlike \cite{MICCAI2016,MICCAI2017}. 

\section{Proposed Approach}
In this work, we employ the convolutional autoencoder which tries to learn compact representative features of the image data. 
The problem to construct HF-like MR images from LF MR images involves the non-linear transformation, which the convolutional autoencoder learns at 
in latent space at multiple scales of the image obtained by upsampling and downsampling layers. The salient aspects of the proposed approach are detailed below: 
\subsection{One encoder with multiple decoders}
For the image reconstruction task, being an ill-posed problem, 
many solutions (HF images) may exist for the transformation of LF image to the HF image estimate. 
The transformation in our case depends on the filter weights which ideally should be representative enough to construct image details of complex structures, and discriminative enough to be able to learn the differences between details of HF and LF image. While, such a transformation can be learnt with a simple convolutional autoencoder (single encoder and decoder), considering that the transformation can be highly non-linear, there could be different weight combinations that can provide good estimates of such a transformation. The proposed multi-decoder model is thoughtfully designed with a notion that decoders initialized randomly, and updated using individual distinct costs, are likely to learn different weights via the different optimization paths. 
The random initializations can yield diverse solutions that can easily be collated for better PSNR. The distinctness between the learnt weights can be shown via activation maps at various decoder layers. 
In Fig.2, we can note such distinctness in the activation maps at same layers of all the three decoders. 


While there can be different configurations of multiple decoders,  as an example, in this work we consider three decoders, integrated with a single encoder in the proposed architecture (Fig. 2). 
In this architecture, a selective backpropagation approach (as elaborated next) (Fig. 1) 
is proposed to enable the weight updates across the three paths, by selecting one decoder out of the three based on their losses, in each weight-update iteration (i.e. for each batch, with multiple batches considered within an epoch). 

\begin{figure*}[ht!]
\centering  \includegraphics[height=140pt,width=320pt]{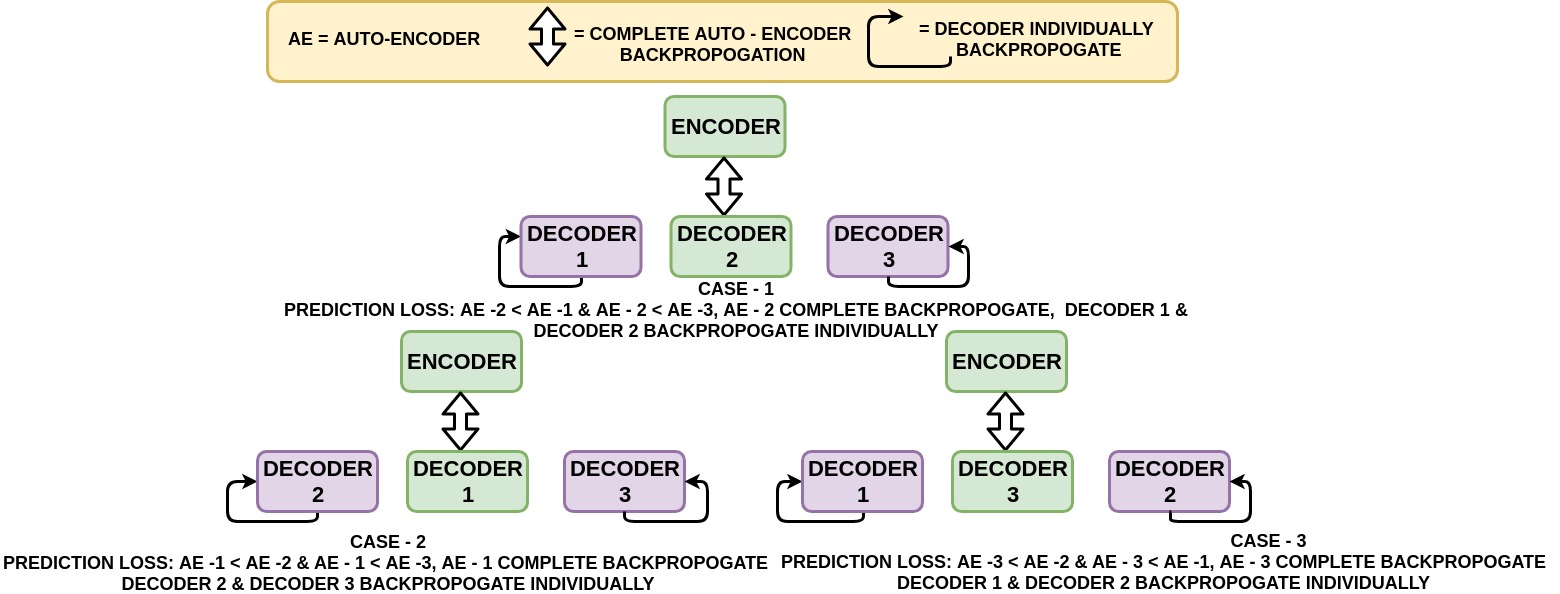}
  \vspace{-0.25cm}
  \caption{\small{Selective Auto-Encoder Backpropagation}}
  	\label{arch}
    \vspace{-0.55cm}
\end{figure*}


\subsection{Updating the weights}
As indicated above, the weights of the architecture are updated in a three-fold manner which involves the selection of one of the 
three decoders, in each iteration. The selection is based on the minimum loss. 

Suppose $\mathbf{E}_i$ represents the error of the network at the $i^{\mbox{th}}$ decoder, such that $\mathbf{E}_i = g(\mathbf{W}_E, \mathbf{W}_{D_i})$, with encoder weights $\mathbf{W}_E$ and decoder weights $\mathbf{W}_{D_i}$, ($i=1, 2, 3.$). The weight update of the encoder is represented as $\Delta\mathbf{W}_E \propto \min_{i}(\mathbf{E}_i)$. In this way, in every iteration the encoder weights have three open but guided paths to move on, and the optimal one (with minimum loss) is chosen. 

While the encoder weights are updated with the minimum decoder loss, for updating the decoder weights, we update all the decoders using their respective losses, i.e. $\Delta\mathbf{W}_{D_i} \propto \mathbf{E}_i$. We observe that 
simultaneously updating the decoders helps in minimizing the training loss faster, even as compared to a single encoder and decoder model, and also yields an improved performance.




 
 
\subsection{Merge Connections}
We define the proposed architecture with blocks of subsequent filter layers followed by a max pooling layer in the encoder section as shown in Fig. 2. To reconstruct the original size of image at output, an upsampling layer is introduced in each block of the decoders. While upsampling, there may be some artifacts introduced due to missing details in downsampled input of decoder. Hence, we concatenate the input of decoder with its upscaled version from the encoder, in order to provide the nature of upscaled details for better reconstruction while upsampling in decoder. Indeed. we observe that adding the merge connections yield a significant PSNR improvement (of the order of 5db).

\subsection{Proposed architecture}
\small
\begin{figure*}[ht!]
\centering \includegraphics[height=200pt,width=355pt]{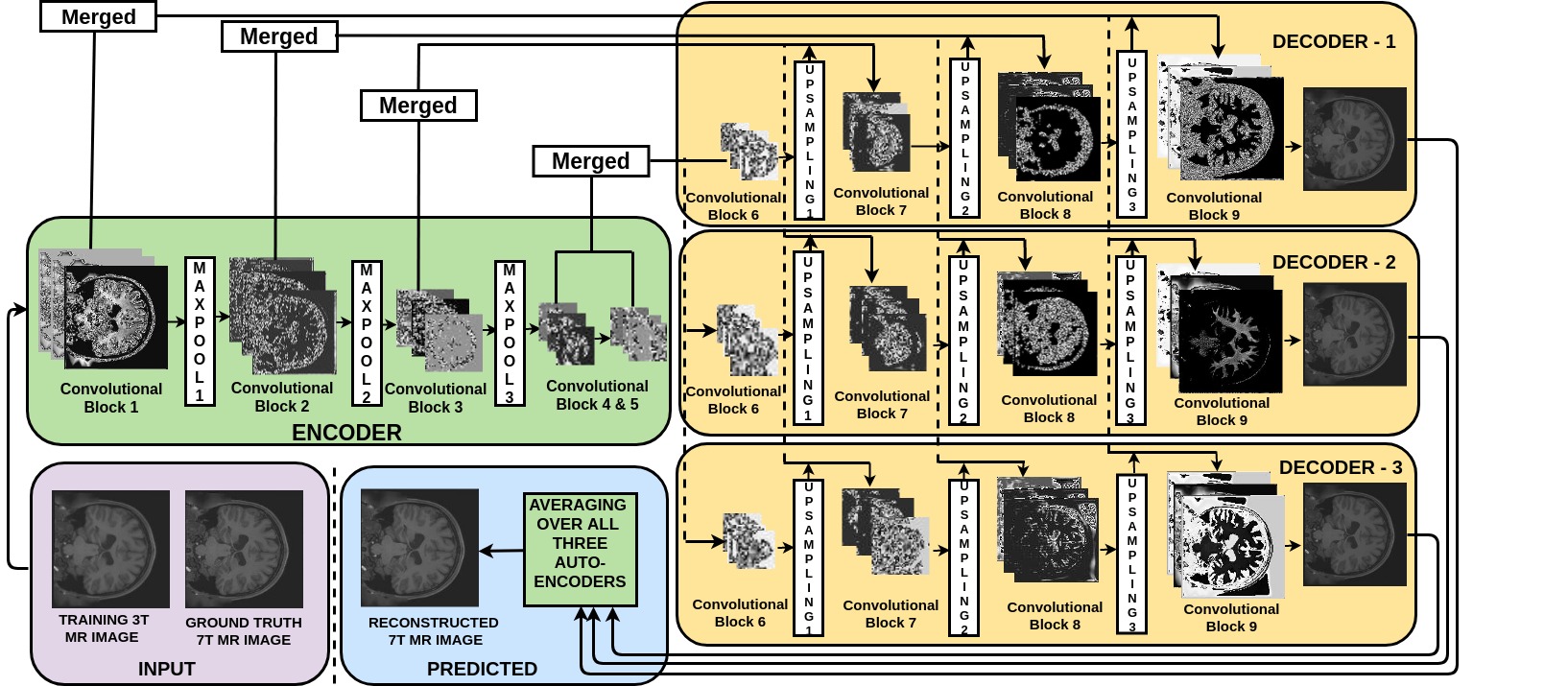}
  \vspace{-0.55cm}
  \caption{\small{Proposed Architecture (Better viewed in color)}}
  	\label{arch}
    \vspace{-0.8cm}
\end{figure*}
\normalsize

The proposed approach employs a single encoder and multiple decoder architecture with a single channel input as described in Fig.\ref{arch}. 
Three convolutional layers are used in each block of an encoder and all the three decoders, followed by a batch normalization layer to maintain the numerical stability. 

The first convolutional block in the encoder has 32 filters and the number of filters doubles after each convolutional block. In all the decoders the first convolutional block has 256 filters and the number of filters are halved after each block. We use a filter size of 3-by-3 in all convolutional blocks. 

We use Rectified Linear Unit (ReLU) as an activation function in all the layers except the final layer. Since our data is normalized between 0 and 1, a Sigmoid activation function is used at the final layer.

The pixel values greater than zero (brain area) are passed to the next layer and rest of the pixels (outer part of the brain) are squashed to zero. This phenomenon is clearly visible in the initial layers of the encoder and the last layers of all the three decoders, through the activation maps in Fig. 2. 

It is well known that local image details at various scales play a significant role in image reconstruction. 
The proposed architecture considers images at different scales using hierarchical layers for downsampling (maxpooling) and upsampling (each for a factor of 2) in encoder and decoders, respectively. 
The encoded representation obtained after three downsampling operations 
brings the data from a high dimension input to a latent space representation.


As the training proceeds, after every epoch the model is validated on 20\% of the unseen validation data.
The auto-encoder with minimum loss on the training data, predicts on the validation data, and using the predicted output we calculate PSNR on the validation data and save the best weights corresponding to the maximum PSNR across epochs. The test reconstructions are then computed on these weights. 


To aid a faster convergence 
we reduce the learning rate by 10\% of its value after every 20 epochs. We also observe that using learning rate decay, the loss value converges to a smaller value than the case without using learning rate decay. The initial value for learning rate is set to be 1e-3. The model is trained for 500 epochs, which is observed to be more than sufficient to ensure convergence. 

Finally, in the testing phase, the three reconstruction estimates on the test data are obtained at the three decoder outputs,  
and We take an average of all the three predictions which improves the results quantitatively.
We observe that averaging the predicted outputs helps in reducing noise effects but preserves the local features in the reconstructed images, due to which the PSNR improves over the individual decoder outputs. 

\section{Experimental Results}
\subsection{Experimental setup}
To evaluate the performance of the proposed algorithm, real MR images scanned by 3T and 7T MR scanner are selected from the dataset available online \cite{dataset}. From a pool of volumes 39 MR image volumes are randomly selected and 3T MR images are registered with 7T MR image volumes using FLIRT software in FSL\cite{FSL}, in order to have pixel to pixel correspondence. Further, each of the MR volume is scaled to 0 to 1 range for numerical stability. The proposed architecture is trained on MR image volumes from 22 subjects, while volumes of 6 subjects are used for validation 
and 11 are used for testing.
We cross validate across 3 trails involving random sets of training, validation and test data. 

For comparison with existing approaches, we re-implement the 3D CNN approach defined in \cite{MICCAI2016}. As some of the parameters are not mentioned in their work, we have used the same parameters as used in proposed work for e.g. learning rate, learning rate decay strategy, optimizer, batch size. To consider a complementary framework, the sparse-representation approach is also used for comparison \cite{YangWright}. For training all approaches we use  207 images from each subject volume. However, due to insignificant information in first and last 20 slices, we select central 167 slices per volume for reconstruction. All implementations are 
on a system with Nvidia 1080 Ti GPU Xeon e5 GeForce processor with 32GB RAM. 

\subsection{Reconstruction results}
The test 7T MR image volumes are constructed using proposed approach and other existing works and two images are randomly selected to illustrate the quality comparison between different approaches. It can be observed from Fig. 3 and Fig. 4, that sparse based approach\cite{YangWright} is able to construct the details but with diffused tissue boundary. 3DCNN performs well in terms of tissue boundary but is unable to restore smaller differences in voxel values. Both these aspects are improved upon by the proposed approach.

The improvement is reflected in the quantitative results in Table 1 with higher PSNR and SSIM values. To compute the performance in terms of blurriness of the edges, two parameters i.e. sharpness and edge width are computed as defined in \cite{juang}. We observe that the algorithm may change the dynamic range of the data. Thus to be consistent for comparing quality of images reconstructed, we first match the histogram (HM) of reconstructed image with the corresponding 3T image. However, we also show the results for the proposed method without HM. The values for parameters are computed over non-background pixels of reconstructed images scaled to their original range.

\begin{figure}[htb]
\small
\begin{minipage}[b]{.10\linewidth}
  \centering
  \centerline{\includegraphics[height=1.493cm,width=1.55cm]{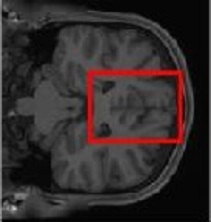}}
 \vspace{0.16cm}
  \centerline{3T Image}\medskip
\end{minipage}\hfill
\begin{minipage}[b]{0.10\linewidth}
  \centering
  \centerline{\includegraphics[trim=14cm 7cm 14cm 2cm,clip=true,height=2cm,width=1.8cm]{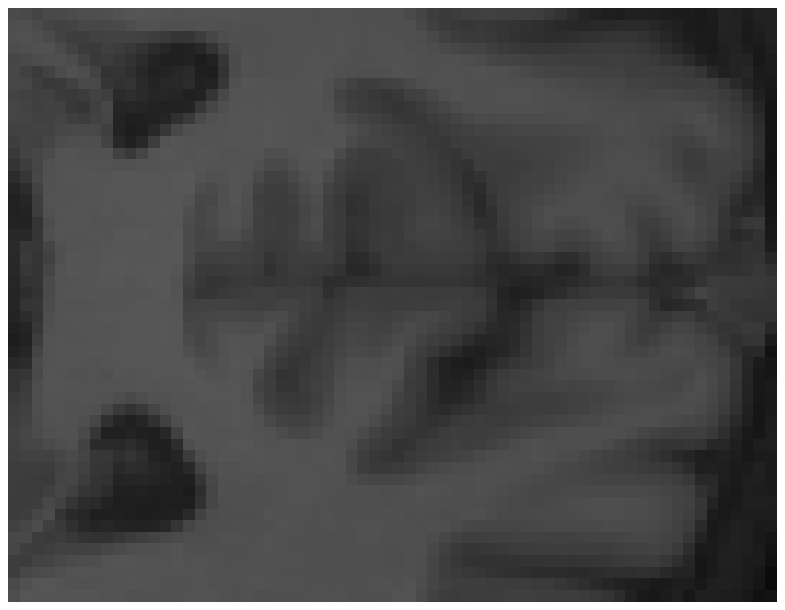}}
  \vspace{0.07cm}
\centerline{Zoomed 3T}\medskip
\end{minipage}
\hfill
\begin{minipage}[b]{0.10\linewidth}
  \centering
  \centerline{\includegraphics[trim=14cm 7cm 14cm 2cm,clip=true,height=2cm,width=1.8cm]{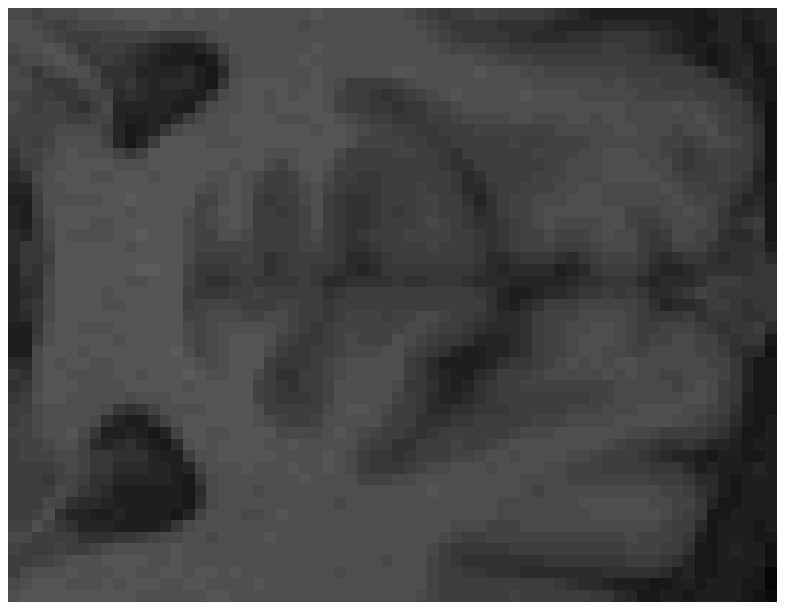}}
  \centerline{ScSR\cite{YangWright}}\medskip
\end{minipage}
\hfill
\begin{minipage}[b]{0.10\linewidth}
  \centering
  \centerline{\includegraphics[trim=14cm 7cm 14cm 2cm,clip=true,height=2cm,width=1.8cm]{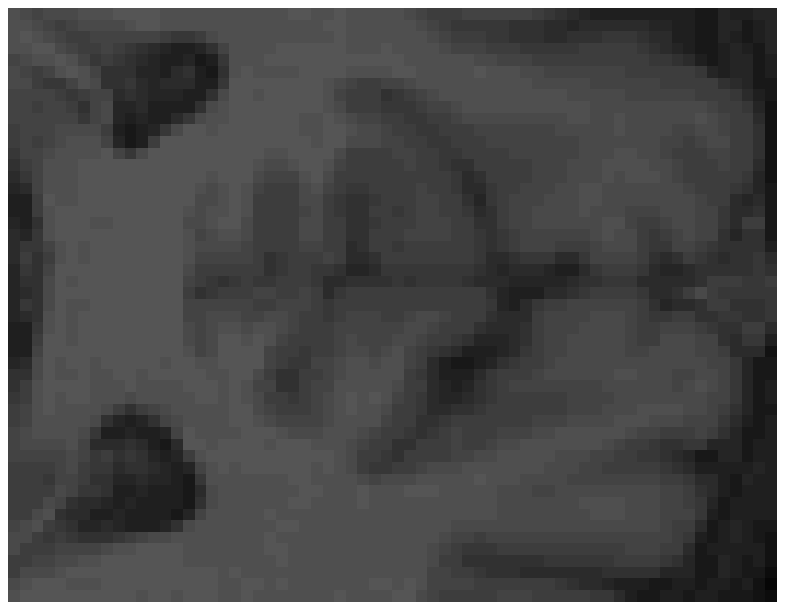}}
  \centerline{3D-CNN\cite{MICCAI2016}}\medskip
\end{minipage}
\hfill
\begin{minipage}[b]{0.10\linewidth}
  \centering
  \centerline{\includegraphics[trim=14cm 7cm 14cm 2cm,clip=true,height=2cm,width=1.8cm]{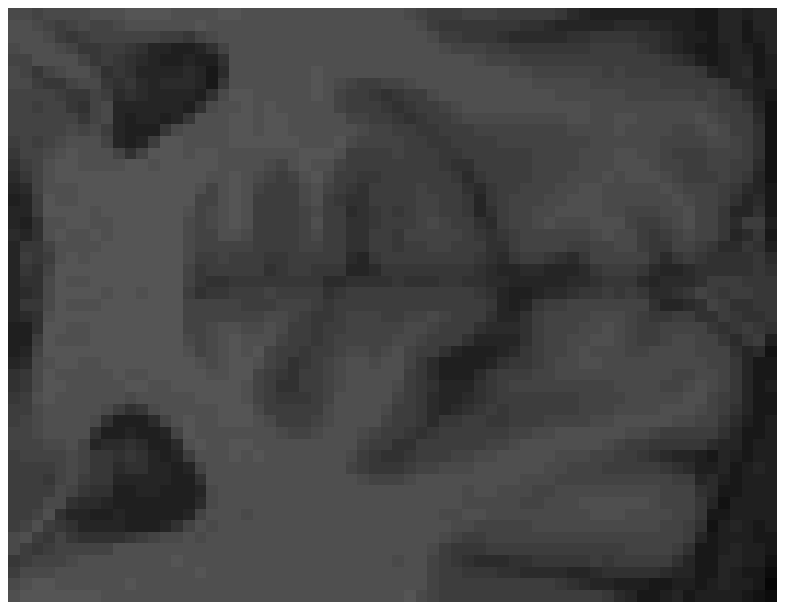}}
  \centerline{Single decoder}\medskip
\end{minipage}
\hfill
\begin{minipage}[b]{0.10\linewidth}
  \centering
  \centerline{\includegraphics[trim=14cm 7cm 14cm 2cm,clip=true,height=2cm,width=1.8cm]{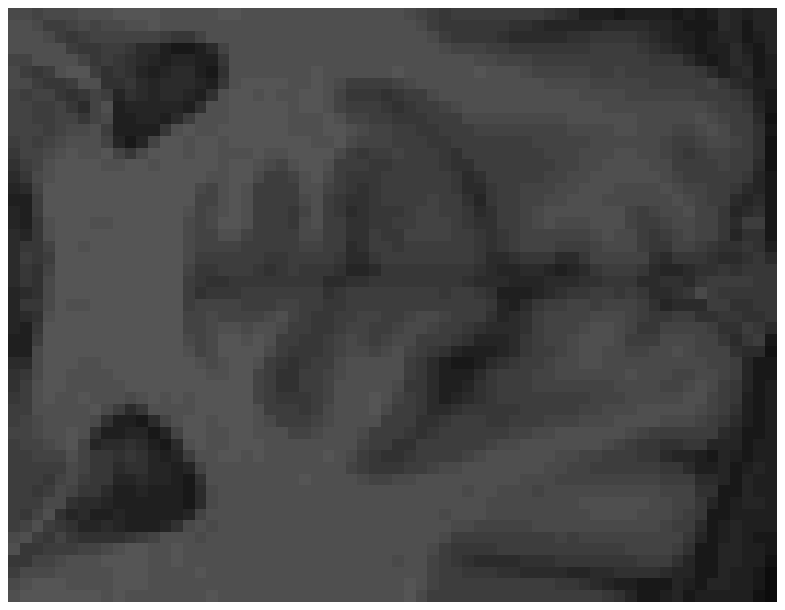}}
  \centerline{Proposed}\medskip
\end{minipage}
\hfill
\begin{minipage}[b]{0.10\linewidth}
  \centering
  \centerline{\includegraphics[trim=14cm 7cm 14cm 2cm,clip=true,height=2cm,width=1.8cm]{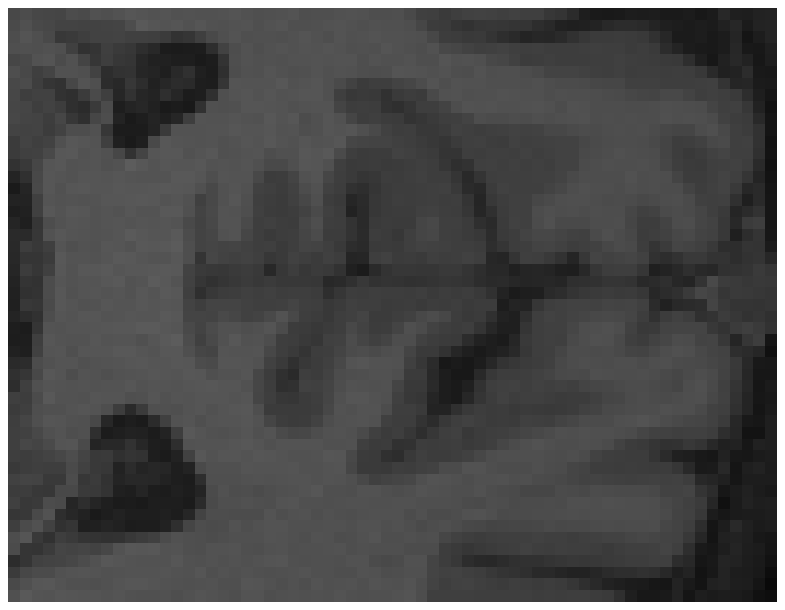}}
  \centerline{7T Image}\medskip
\end{minipage}
 \vspace{-0.28cm}
\small{\caption{\small{Example reconstructions and comparison visualized at a finer scale.} }}
\label{QualityMatters}
\vspace{-0.3cm}
\end{figure}

\begin{figure}[htb]
\small
\begin{minipage}[b]{.10\linewidth}
  \centering
  \centerline{\includegraphics[height=1.493cm,width=1.6cm]{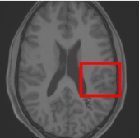}}
 \vspace{0.0cm}
  \centerline{3T Image}\medskip
\end{minipage}\hfill
\begin{minipage}[b]{0.10\linewidth}
  \centering
  \centerline{\includegraphics[height=1.48cm,width=1.7cm]{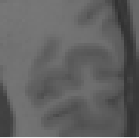}}
  \vspace{0.07cm}
\centerline{Zoomed 3T}\medskip
\end{minipage}
\hfill
\begin{minipage}[b]{0.10\linewidth}
  \centering
  \centerline{\includegraphics[height=1.47cm,width=1.7cm]{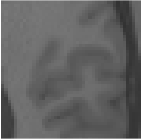}}
  \centerline{ScSR\cite{YangWright}}\medskip
\end{minipage}
\hfill
\begin{minipage}[b]{0.10\linewidth}
  \centering
  \centerline{\includegraphics[height=1.48cm,width=1.7cm]{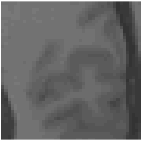}}
  \centerline{3D-CNN\cite{MICCAI2016}}\medskip
\end{minipage}
\hfill
\begin{minipage}[b]{0.10\linewidth}
  \centering
  \centerline{\includegraphics[height=1.493cm,width=1.8cm]{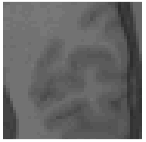}}
  \centerline{Single decoder}\medskip
\end{minipage}
\hfill
\begin{minipage}[b]{0.10\linewidth}
  \centering
  \centerline{\includegraphics[height=1.49cm,width=1.75cm]{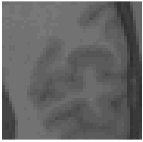}}
  \centerline{Proposed}\medskip
\end{minipage}
\hfill
\begin{minipage}[b]{0.10\linewidth}
  \centering
  \centerline{\includegraphics[height=1.498cm,width=1.75cm]{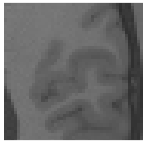}}
  \centerline{7T Image}\medskip
\end{minipage}
\vspace{-0.4cm}
\small{\caption{\small{Example reconstructions and comparison visualized at a finer scale.} }}
\label{QualityMatters}
\vspace{-0.55cm}
\end{figure}

\subsection{Segmentation results}
High quality images helps in improving segmentation of the tissues required for medical analysis. Thus, we compare segmentation labels for images reconstructed by different algorithms, with FAST software of FSL for gray matter(GM), white matter(WM) and CSF. 
The dice-ratio improvements in segmentation with reconstruction using the proposed approach is clear from  
Table 2. 
The work in \cite{MICCAI2016} has outperformed the sparse based reconstruction, thus we do not provide segmentation results for the latter. 


%
%

%

\vspace{-0.4cm}
\begin{table}[H] 
    \label{table}
    \small
    \begin{center}
    \begin{tabular}
    {|p{2.4cm}|p{1.5cm}|p{1.5cm}|p{2.4cm}|p{1.7cm}|p{1.6cm}|}
    \hline
\textbf{Approaches} & \textbf{ScSR with HM\cite{YangWright}} & \textbf{3D-CNN with HM\cite{MICCAI2016}} & \textbf{Single decoder with HM} & \textbf{Proposed Approach with HM} & \textbf{Proposed Approach w/o HM}
    \\ \hline
    \textbf{PSNR (dB)} &
 {35.96 \newline ($\pm$0.93)} & {34.20}\newline ($\pm$0.81) & {{36.96}}\newline ($\pm$0.92) & {37.45}\newline ($\pm$1.00) & {39.25}\newline ($\pm$1.46)
    \\ \hline
 \textbf{Average SSIM} &
 {0.7092} & {0.7094} & {0.7371} & {0.7432	} & {0.8253}
\\ \hline
\textbf{Sharpness\cite{juang}} &
 {0.3967} & {0.4043} & {0.4092} & {0.4179} & {0.4001}
\\ \hline
	\textbf{Edge width\cite{juang}} &
 {0.0989} & {0.0945} & {0.0947} & {0.0919} & {0.0959}
\\ \hline
    \end{tabular}
    \end{center}
    \caption{\small{Quantitative comparison of proposed approach}}
    \vspace{-1.0cm}
 \end{table}

\vspace{-0.4cm}
\begin{table}[H]
    \label{table}
    \small
    \begin{center}
    \begin{tabular}{|p{2.6cm}|p{2.5cm}|p{2.5cm}|p{2.5cm}|}
    \hline
\textbf{Approaches} & \textbf{3T MR Images} & \textbf{3D-CNN with HM\cite{MICCAI2016}} & \textbf{Proposed approach}
    \\ \hline
    \textbf{CSF} &
 {0.8836 ($\pm$0.0081)} & {0.8766 ($\pm$0.0053)} & {0.9149 ($\pm$0.0042)} \\ \hline
 \textbf{White Matter} &
 {0.9372 ($\pm$0.0086)} & {0.9279 ($\pm$0.0100)} & {0.9528 ($\pm$0.0068)}
\\ \hline
\textbf{Gray Matter} &
 {0.9503  ($\pm$0.0083)} & {0.9216 ($\pm$0.0157)} &{0.9602 ($\pm$0.0087)}
\\ \hline
    \end{tabular}
    \end{center}
    \caption{\small{Dice ratio for segmentation of images reconstructed by different algorithms}}
    \vspace{-0.6cm}
\end{table}

\vspace{-0.5cm}
\subsection{Computational Complexity}
Here, we stress the computational advantage of the proposed approach in terms of run-time for reconstruction, as compared to the approach of \cite{MICCAI2016}. 
The 3D CNN approach \cite{MICCAI2016} takes 137 minutes to construct 11 subject image volume. On the other hand, the proposed algorithm is computationally simple and takes less than 2 minutes to do the same task.
To justify this advantage over [5], we note that the amount of multiplications in the architecture of [5] is 2145 times than that in the proposed one. This is largely due to 3D convolution in [5] vs 2D in ours, and unpadded convolution in [5] vs max-pooling in ours (latter yields more dimensionality reduction).

\section{Conclusion}
We reported a novel convolutional single encoder with three decoder framework for reconstructing 7T-like MR images from 3T MR image as inputs. The proposed approach employs single-channel input (i.e. does not require anatomical and segmentation features as an input), and yet achieves a superior reconstruction quality 
over some contemporary methods. It also has a significant computational advantage. We also show that the reconstructed 7T-like MR images when segmented have better dice ratio compared to the comparative approaches.
\vspace{-0.2cm}
\small
\bibliographystyle{IEEEtran}
\bibliography{refers.bib}

\begin{thebibliography}{1}
\providecommand{\url}[1]{#1}
\csname url@samestyle\endcsname
\providecommand{\newblock}{\relax}
\providecommand{\bibinfo}[2]{#2}
\providecommand{\BIBentrySTDinterwordspacing}{\spaceskip=0pt\relax}
\providecommand{\BIBentryALTinterwordstretchfactor}{4}
\providecommand{\BIBentryALTinterwordspacing}{\spaceskip=\fontdimen2\font plus
\BIBentryALTinterwordstretchfactor\fontdimen3\font minus
  \fontdimen4\font\relax}
\providecommand{\BIBforeignlanguage}[2]{{%
\expandafter\ifx\csname l@#1\endcsname\relax
\typeout{** WARNING: IEEEtran.bst: No hyphenation pattern has been}%
\typeout{** loaded for the language `#1'. Using the pattern for}%
\typeout{** the default language instead.}%
\else
\language=\csname l@#1\endcsname
\fi
#2}}
\providecommand{\BIBdecl}{\relax}
\BIBdecl

\bibitem{CanThey}
\BIBentryALTinterwordspacing
E.~Plenge, D.~H.~J. Poot, M.~Bernsen, G.~Kotek, G.~Houston, P.~Wielopolski,
  L.~van~der Weerd, W.~J. Niessen, and E.~Meijering, ``Super-resolution methods
  in mri: Can they improve the trade-off between resolution, signal-to-noise
  ratio, and acquisition time?'' \emph{Magnetic Resonance in Medicine},
  vol.~68, no.~6, pp. 1983--1993, 2012. [Online]. Available:
  \url{http://dx.doi.org/10.1002/mrm.24187}
\BIBentrySTDinterwordspacing

\bibitem{TMI3T7T}
K.~Bahrami, F.~Shi, X.~Zong, H.~W. Shin, H.~An, and D.~Shen, ``Reconstruction
  of 7t-like images from 3t mri,'' \emph{IEEE Transactions on Medical Imaging},
  vol.~35, no.~9, pp. 2085--2097, Sept 2016.

\bibitem{MIMCS}
\BIBentryALTinterwordspacing
S.~Roy, A.~Carass, and J.~L. Prince, ``Magnetic resonance image example based
  contrast synthesis,'' \emph{IEEE Trans Med Imaging}, vol.~32, no.~12, pp.
  2348--2363, Dec 2013. [Online]. Available:
  \url{http://www.ncbi.nlm.nih.gov/pmc/articles/PMC3955746/}
\BIBentrySTDinterwordspacing

\bibitem{YangWright}
J.~Yang, J.~Wright, T.~S. Huang, and Y.~Ma, ``Image super-resolution via sparse
  representation,'' \emph{IEEE Transactions on Image Processing}, vol.~19,
  no.~11, pp. 2861--2873, Nov 2010.

\bibitem{MICCAI2016}
K.~Bahrami, F.~Shi, I.~Rekik, and D.~Shen, ``Convolutional neural network for
  reconstruction of 7t-like images from 3t mri using appearance and anatomical
  features,'' in \emph{Deep Learning and Data Labeling for Medical
  Applications}, 2016, pp. 39--47.

\bibitem{MICCAI2017}
K.~Bahrami, I.~Rekik, F.~Shi, and D.~Shen, ``Joint reconstruction and
  segmentation of7t-like mr images from 3t mri based oncascaded convolutional
  neural networks,'' in \emph{Medical Image Computing and Computer Assisted
  Intervention − MICCAI 2017}, 2017, pp. 764--772.

\bibitem{dataset}
``https://www.humanconnectome.org/study/hcp-young-adult/document/1200-subjects-data-release.''

\bibitem{FSL}
\BIBentryALTinterwordspacing
F.~Shi, L.~Wang, Y.~Dai, J.~H. Gilmore, W.~Lin, and D.~Shen, ``Label: Pediatric
  brain extraction using learning-based meta-algorithm,'' \emph{NeuroImage},
  vol.~62, no.~3, pp. 1975 -- 1986, 2012. [Online]. Available:
  \url{http://www.sciencedirect.com/science/article/pii/S1053811912005307}
\BIBentrySTDinterwordspacing

\bibitem{juang}
\BIBentryALTinterwordspacing
J.~Guan, W.~Zhang, J.~Gu, and H.~Ren, ``No-reference blur assessment based on
  edge modeling,'' \emph{Journal of Visual Communication and Image
  Representation}, vol.~29, pp. 1 -- 7, 2015. [Online]. Available:
  \url{http://www.sciencedirect.com/science/article/pii/S1047320315000085}
\BIBentrySTDinterwordspacing

\end{thebibliography}

\end{document}